\documentclass[runningheads]{llncs}
\usepackage{graphicx}
\usepackage{comment}
\usepackage{amsmath,amssymb} 
\usepackage{color}
\usepackage{booktabs}
\usepackage{cite}
\usepackage{url}
\usepackage{hyperref}
\hypersetup{
	colorlinks=true,
	linkcolor=blue,
	filecolor=magenta,
	urlcolor=magenta,
}

\begin{document}
\pagestyle{headings}
\mainmatter
\def\ECCVSubNumber{2897}  

\title{3D Bird Reconstruction: a Dataset, Model, and Shape Recovery from a Single View} 

\titlerunning{3D Bird Reconstruction}
%
\author{Marc Badger\inst{1,2} \and
Yufu Wang\inst{1,2} \and
Adarsh Modh\inst{1,2} \and
Ammon Perkes\inst{1,2} \and
Nikos Kolotouros\inst{1,2} \and
Bernd G. Pfrommer\inst{1,2} \and
Marc F. Schmidt\inst{1,3} \and
Kostas Daniilidis\inst{1,2}}

\authorrunning{M. Badger et al.}
%
\institute{University of Pennsylvania, Philadelphia PA 19104, USA \and
\email{\{mbadger, yufu, adarshm, nkolot, pfrommer, kostas\}@seas.upenn.edu} \and
\email{\{aperkes, marcschm\}@sas.upenn.edu}}
\maketitle

\begin{abstract}
Automated capture of animal pose is transforming how we study neuroscience and social behavior. Movements carry important social cues, but current methods are not able to robustly estimate pose and shape of animals, particularly for social animals such as birds, which are often occluded by each other and objects in the environment. To address this problem, we first introduce a model and multi-view optimization approach, which we use to capture the unique shape and pose space displayed by live birds. We then introduce a pipeline and experiments for keypoint, mask, pose, and shape regression that recovers accurate avian postures from single views. Finally, we provide extensive multi-view keypoint and mask annotations collected from a group of 15 social birds housed together in an outdoor aviary. The project website with videos, results, code, mesh model, and the Penn Aviary Dataset can be found at \href{https://marcbadger.github.io/avian-mesh}{https://marcbadger.github.io/avian-mesh}.
\end{abstract}

\section{Introduction}

\textbf{Why computational ethology?} Accurate measurement of behavior is vital to disciplines ranging from neuroscience and biomechanics to human health and agriculture. Through automated measurement, computational ethology aims to capture complex variation in posture, orientation, and position of multiple individuals over time as they interact with each other and their environment \cite{Anderson2014}. Pose trajectories contain rich, unbiased, information from which we can extract more abstract features that are relevant to brain function, social interactions, biomechanics, and health. Studying neural functions in the context of natural social behavior is a critical step toward a deeper understanding how the brain integrates perception, cognition, and learning and memory to produce behavior. Pose trajectories reveal how animals maneuver to negotiate cluttered environments, how animals make decisions while foraging or searching for mates, and how the collective behavior of a group arises from individual decisions. Automated capture of difficult-to-observe behaviors is transforming diverse applications by streamlining the process of extracting quantitative physiological, behavioral, and social data from images and video.

\begin{figure}
\begin{center}
\includegraphics[width=0.78\linewidth]{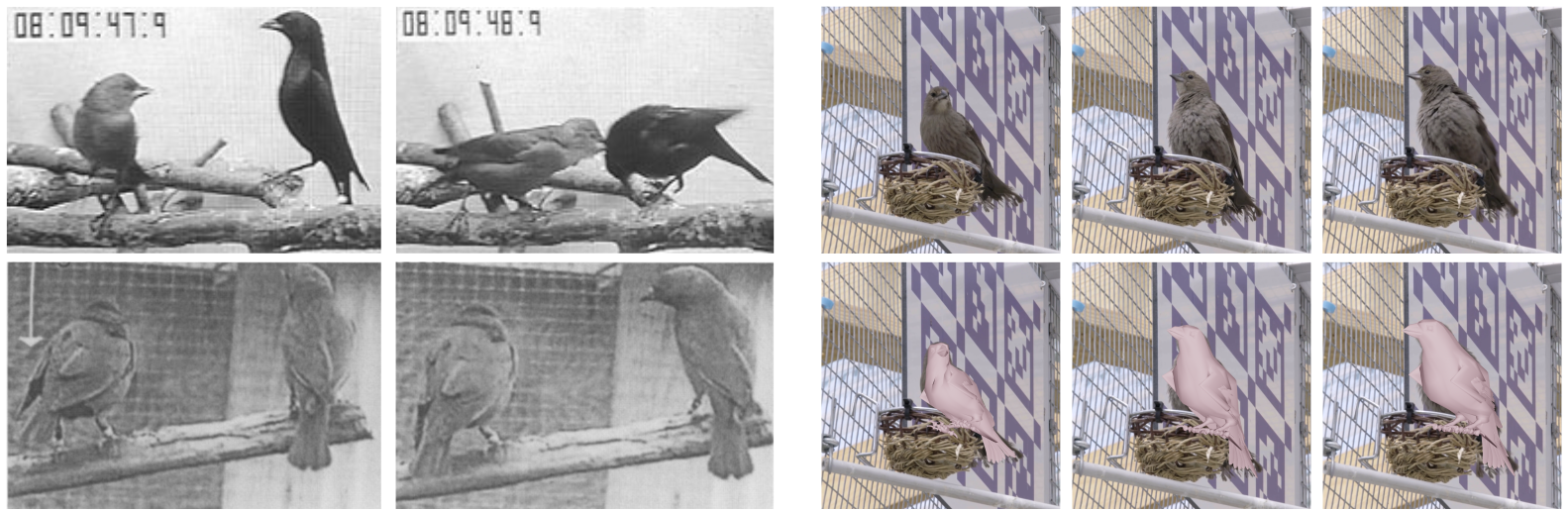}
\end{center}
    \caption{Visual signals convey important social cues in birds. Motions such as pecking (top left) and wingstrokes (bottom left) drive social behavior in both males and females. Complex wing folding and large changes in body volume when feathers are puffed (upper right) make shape recovery (lower right) a difficult task. Images from \cite{West1988} and \cite{Baillie2019}.}
\label{fig:wingstrokes}
\end{figure}

\textbf{Why does bird posture matter? Why cowbirds?} Understanding how the collective behavior of social groups arises from individual interactions is important for studying the evolution of sociality and neural mechanisms behind social behaviors. Although vocalizations are a clear channel for communication in birds, surprisingly, changes in posture, orientation, and position also play an important role in communication. One of the best studied groups from both behavioal and neuroscience perspectives are the brown-headed cowbirds (\textit{Molothrus ater}). \textbf{In cowbirds, females influence the behavior of males through a number of visual mechanisms including ``wingstrokes'', which involve changes in both pose and shape over time} \cite{West1988} (Figure \ref{fig:wingstrokes}). Interactions between birds are usually recorded by observing a focal individual's interactions in person in the field. Although insightful, such manual observations contain observer bias, miss interactions between non-focal individuals, and cannot be performed continuously for long periods. Qualitative observations also miss important variation in posture that would be revealed by a quantitative approach. For example, Figure \ref{fig:wingstrokes} shows changes in pose and shape that can serve as social cues in cowbirds. The ability to estimate the pose of multiple interacting individuals would transform the study of animal communication \cite{Anderson2014}, as is it beginning to do for humans \cite{Pavlakos2019X, Joo2015, Joo2019}. Estimating the pose and shape of birds in a social context, however, presents several challenges.
\begin{figure}
\begin{center}
\includegraphics[width=0.95\linewidth]{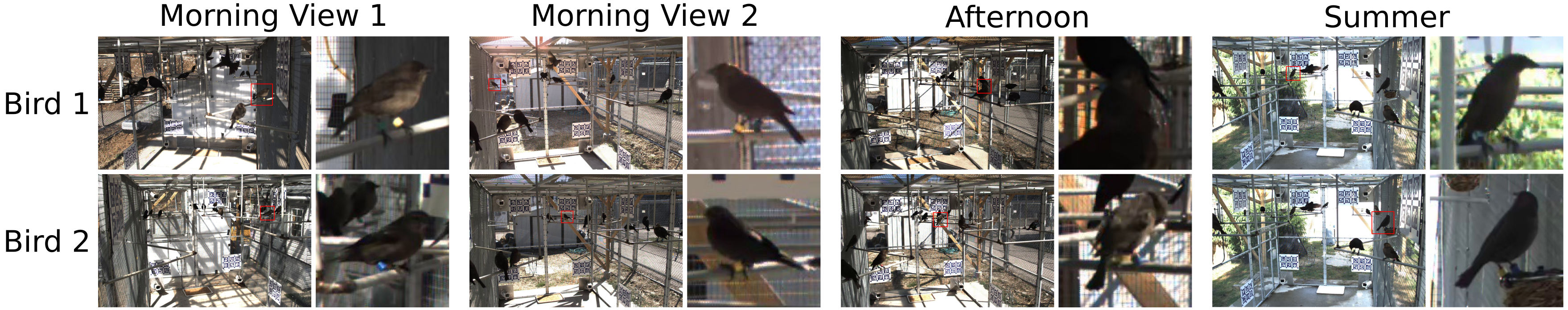}
\end{center}
    \caption{Appearance variation across bird identity (top vs bottom) and across viewpoint, time of day, and season (1st column vs. columns 2-4 respectively). The red box within the left image of each panel shows the location of the enlarged crop (right image).}
\label{fig:variation}
\end{figure}

\textbf{Why is estimating bird pose and shape challenging?} Recovering shape and pose of birds in the wild is challenging for the following four reasons: 
\begin{enumerate}
\item Changes in pose and shape are difficult to model in birds.
\item No pose or shape priors are available.
\item Many birds are only visible from a single unoccluded view.
\item Appearance variation in natural settings makes detection difficult.
\end{enumerate}
Shape is particularly difficult to model because birds have highly mobile feathers that allow dramatic changes in both shape (e.g. tail fanning) and perceived body volume (e.g. feather puffing in Figure \ref{fig:wingstrokes}).  Furthermore, when the wings are held next to the body, they are folded in a complex way in which much of the wing surface becomes sandwiched between the top of the wing and the body.  These ``internal'' surfaces cannot be recovered from scans of figurines with folded wings and figurines with wings in intermediate poses are not available. In addition to modeling challenges, cowbirds interact in a complex environment containing extreme variation in illumination and may be heavily occluded either by vegetation by other birds in the group.

Animal posture is often described using joint angles derived from semantic keypoints, joints, or other anatomical locations.  This approach is attractive because keypoints are easy to identify and can readily be localized by deep-learning-based software packages such as DeepLabCut~\cite{Mathis2018}, DeepPoseKit~\cite{Graving2019}, and LEAP~\cite{Pereira2019}. Under heavy occlusion, however, even multi-view setups frequently do not observe relevant keypoints from more than one view. One solution is to lift the pose from 2D to 3D, but unlike for humans, researchers do not yet have strong species-specific priors for tackling this problem. We overcome the limitations of directly using 2D keypoints and skeletons by fitting a 3D parameterized mesh model with priors learned from a multi-view dataset.

\textbf{Dataset.} With the aim of creating a robust system for estimating the shape and pose of multiple interacting birds over months-long timescales, we recorded the behavior of 15 cowbirds housed together in an outdoor aviary over the course of a single three-month mating season. Our carefully calibrated multi-view dataset contains large variation in (i) bird pose, orientation, and position/depth, (ii) viewpoint across eight cameras, and (iii) appearance across different lighting conditions (time of day and weather) and seasons (Figure \ref{fig:variation}). Cowbirds have a nearly textureless appearance and birds move freely and explore all three dimensions of their cage, producing a large range of subject depth with respect to the camera.  Importantly, both perched and flying birds adopt postures covering a large range of motion in both orientation and pose.

\begin{figure}
\begin{center}
    \includegraphics[width=0.95\linewidth]{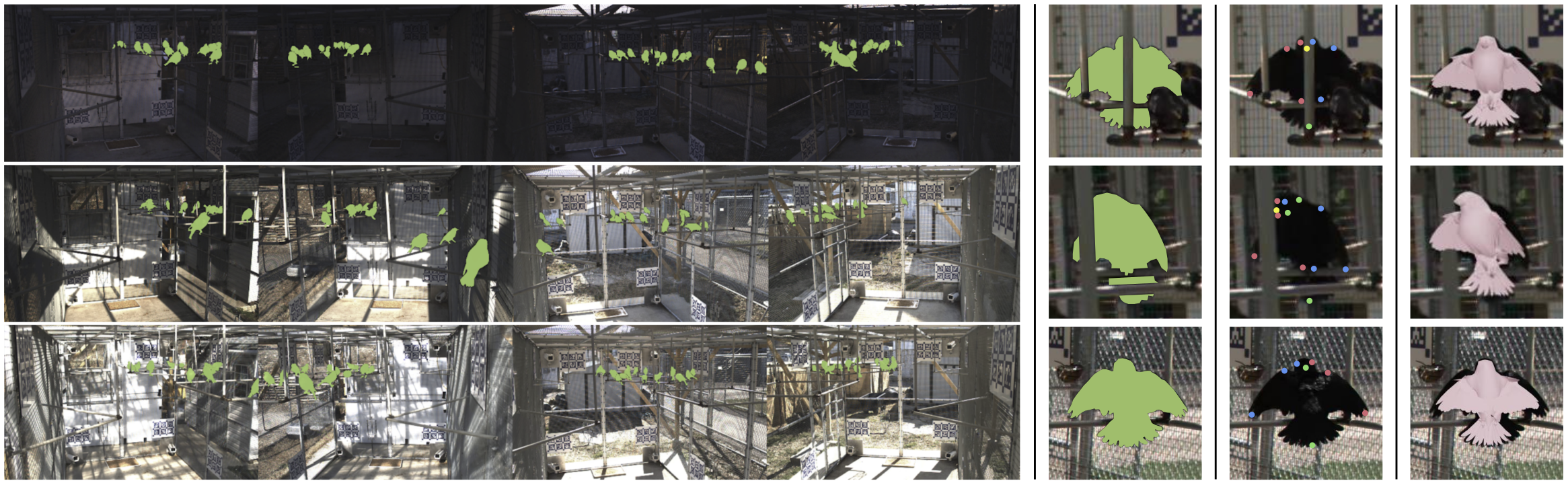}
\end{center}
    \caption{The dataset and model. We provide multi-view segmentation masks for over 6300 bird instances, keypoints for 1000 bird instances, the first articulated 3D mesh model of a bird, and a full pipeline for recovering the shape and pose of birds from single views.}
\label{fig:teaser}
\end{figure}

We annotated silhouette and keypoints for 1000 instances and matched these annotations across views. Although 90\% of annotated birds were visible from 3 or more cameras, about half of the annotated instances were occluded to some degree. Only 62\% of birds had more than one unoccluded view, highlighting the need for a single-view approach.

\begin{figure}
\begin{center}
\includegraphics[width=0.9\linewidth]{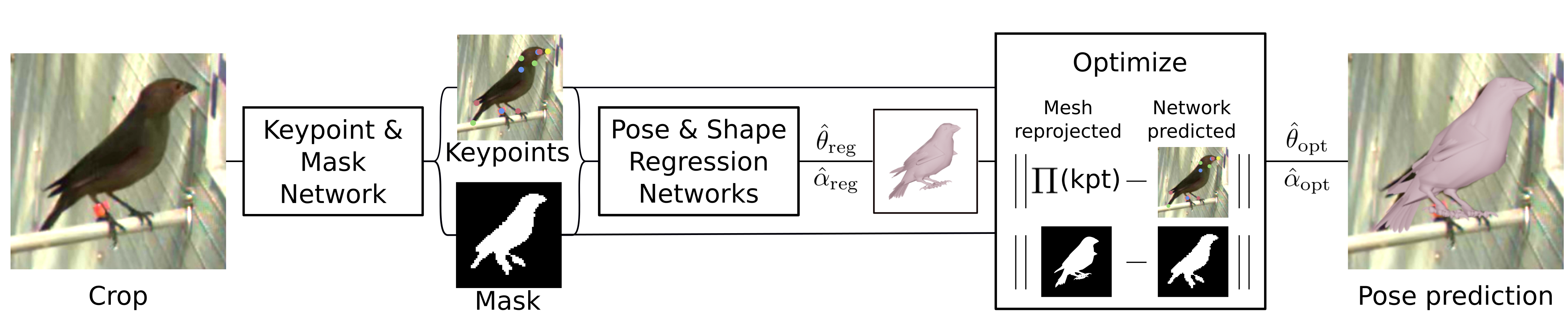}
\end{center}
   \caption{We estimate the 3D pose and shape of birds from a single view. Given a detection and associated bounding box, we predict body keypoints and a mask.  We then predict the parameters of an articulated avian mesh model, which provides a good initial estimate for optional further optimization.}
\label{fig:fullpipeline}
\end{figure}

After collecting keypoint and silhouette ground truth from multiple views, we fit our avian mesh model using a multi-view optimization-based approach to learn a shape space and pose priors.  We then use the model and priors to train a neural network to regress pose parameters directly from keypoint and silhouette data. These pose parameters can be used to initialize a single-view optimization procedure to further refine body pose and shape (Figure \ref{fig:fullpipeline}).  We use our dataset for learning instance segmentation and keypoint localization, and for estimating bird pose and shape, but our dataset could also be used in the future for learning Re-ID tasks.

In summary, our contributions are focused around the four challenges mentioned previously:
\begin{enumerate}
\item We develop the first parameterized avian mesh model that is capable of capturing the unique pose and shape changes displayed by birds.
\item We fit our mesh model to available multi-view keypoint and silhouette data using an optimization-based approach to obtain an accurate shape space and pose prior.
\item We develop a neural network based pipeline for recovering the shape and pose of birds from a single view.
\item We present a challenging multi-view dataset for studying social behavior in birds. The dataset contains extreme variation in subject appearance and depth and many subjects are fully or partially occluded in all but one view.
\end{enumerate}

\section{Related work}

\textbf{Human pose and shape estimation.}
Recent advances in human pose estimation have capitalized on i) powerful 2D joint detectors, ii) 3D pose priors, and iii) low-dimensional articulated 3D shape models of the human body.  SMPL~\cite{Loper2015}, the most popular formulation, first deforms a template mesh using shape and pose parameters learned from over 1000 registered body scans of people~\cite{Bogo2014} and then uses linear blend skinning (LBS) to transform mesh vertices given a set of joint angles.  In SMPLify, Bogo et al.~\cite{Bogo2016} estimate 3D human pose and shape from single images by fitting SMPL to 2D keypoints.  Huang et al.~\cite{Huang2017} extend SMPLify~\cite{Bogo2016} to the multi-view setting and show a positive effect of silhouette supervision in addition to keypoints. Pavlakos et al.~\cite{Pavlakos2018} estimate pose and shape directly from predicted keypoints and silhouettes in an end-to-end framework. Recent approaches regress pose and shape directly from images and use adversaries with access to a 3D pose dataset~\cite{Kanazawa2018hmr}, Graph-CNN architectures~\cite{Kolotouros2019CMR}, texture consistency~\cite{Pavlakos2019Texture}, and model-fitting within the training loop~\cite{Kolotouros2019SPIN}. All of the above methods base their approach on parameterized mesh models indicating their critical importance for bridging between observation in 2D and estimation in 3D.  In contrast to previous works that rely on 3D scans and SMPL-like models to develop meshes and shape spaces for novel domains such as hands \cite{Romero2017}, faces \cite{Li2017}, and four-legged animals \cite{Zuffi2017}, we learn our avian mesh model directly from video data of live birds.

\textbf{Animal pose and shape estimation.}
Within biology, most work focuses on isolated animals with no background clutter and few occlusions. Mathis et al. \cite{Mathis2018} and Pereira et al. \cite{Pereira2019} recently provided tools for training convolutional neural networks for keypoint localization. Graving et al. \cite{Graving2019} localize keypoints on three datasets of fruit flies \cite{Pereira2019}, desert locusts \cite{Graving2019}, and Gr\'evy's zebras \cite{Graving2019}. G\"unel et al. \cite{gunel2019} use a Stacked Hourglass network \cite{Newell2016} for 2D keypoint localization in flies and perform pictorial structures and belief propagation message passing \cite{Felzenszwalb2005} to reconstruct 3D pose from 2D detections. Liu and Belhumeur et al.~\cite{Liu2013} use HOG descriptors and linear SVMs to localize bird parts in the more challenging CUB-200-2011 dataset~\cite{Wah2011}. All of these works are based on the detection and direct triangulation of 2D keypoints.  A fundamental challenge, however, is that any particular keypoint may not be visible from more than one view.  Models that constrain the relative position of keypoints, such as the parameterized mesh model we present here, overcome this issue.

Two previous works use articulated graphics models to estimate the pose of flying animals. Fontaine et al.~\cite{Fontaine2009} construct a 3D mesh model of a fruit fly and estimate the fly's trajectory and pose over time by fitting the model to three orthogonal views. Breslav~\cite{Breslav2016} create a two-DOF 3D graphics model of a bat and use a Markov Random Field to estimate the 3D pose of bats flying in the wild captured with a multi-view thermal camera setup.

Animal shape estimation is a difficult task. Cashman and Fitzgibbon~\cite{Cashman2013} estimate the shape of dolphins.  Ntouskos et al.~\cite{Ntouskos2015} fit shape primitives to silhouettes of four legged animals.  Vincente and Agapito~\cite{Vincente2013} obtain and deform a template mesh using silhouettes from two reference images.  Kanazawa et al.~\cite{Kanazawa2016} learn how animals deform from images by creating an animal-specific model of local stiffness. Kanazawa et al.~\cite{Kanazawa2018cmr} predict shape, pose, and texture of birds in CUB-200 by deforming a spherical mesh, but do not model pose and thus the locations of wingtips on the mesh are often topologically adjacent to the tail rather than near the shoulders. Zuffi et al.~\cite{Zuffi2017} create a realistic, parameterized 3D model (SMAL) from scans of toys by aligning a four-legged template to the scans. They capture shape using PCA coefficients of the aligned meshes and learn a pose prior from a short walking video. Zuffi, Kanazawa, and Black~\cite{Zuffi2018} fit the SMAL model to several images of the same animal and then refine the shape to better fit the image data, resulting in capture of both shape and texture (SMALR). Zuffi et al.~\cite{Zuffi2019} estimate 3D pose, shape, and texture of zebras in the wild by integrating the SMAL model into an end-to-end network regression pipeline.  Their key insight was to first use SMALR to pose an existing horse model and capture a rough texture of the target species.  A common feature of these approaches is that they create or leverage a parameterized mesh model.  The SMAL model was only trained on four-legged animals so the shape space learned by the model is insufficient for modeling birds, which differ markedly in both limb shape and joint angles. To overcome the lack of a statistical model for birds, we add one additional degree of freedom to each joint and obtain a pose and shape space from multi-view fits to live birds.

\textbf{Datasets for animal pose estimation.}
Large-scale object recognition datasets contain many species of animals including dogs, cats, birds, horses, sheep, and more. MS COCO \cite{cocodataset} contains 3362 images with bird mask annotations, but no keypoint or pose annotations. The CUB-200 dataset \cite{Wah2011} contains 11,788 masks and keypoint instances of birds in the wild. A fruit fly dataset \cite{Pereira2019} contains 1500 images with centered dorsal views of single flies walking in an arena with a plain white background containing no variation or distractors. The desert locust (800 images) and Gr\'evy's zebras (900 images) include other individuals in the frame, but views are dorsal-only, centered, and narrowly cropped around a focal individual. In contrast our multi-view dataset contains both masks and keypoints of multiple, overlapping subjects and has large variation in relative viewpoint and complex changes in background and lighting.

\section{Approach}
\begin{figure}
\begin{center}
\includegraphics[width=0.9\linewidth]{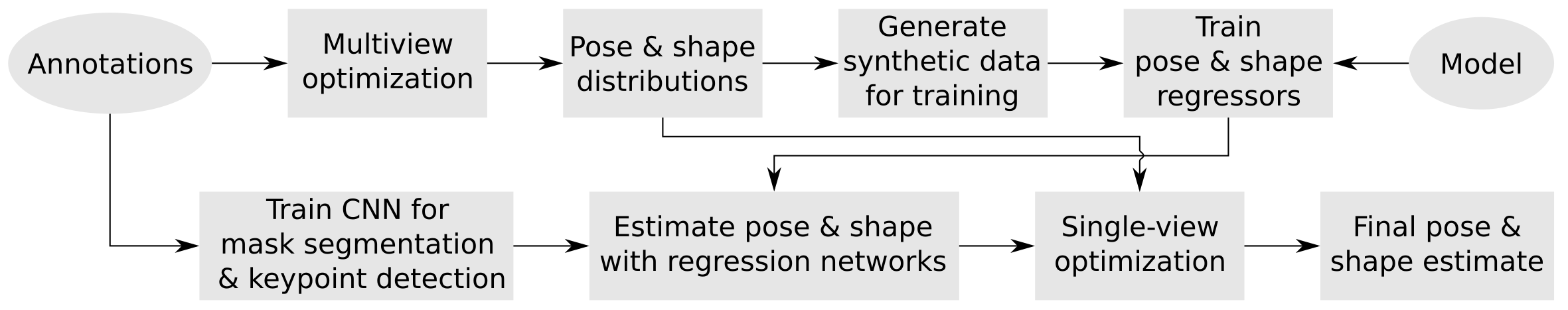}
\end{center}
   \caption{Overall approach for recovering bird pose and shape from a single view. See Figure \ref{fig:fullpipeline} for a detailed view of the final pipeline.}
\label{fig:approach}
\end{figure}

We use a boot-strapped, four-step approach to developing a full pipeline for 3D bird reconstruction from single images (Figure \ref{fig:approach}). First we develop a parameterized avian mesh and use a multi-view optimization procedure to fit the model to annotations in our dataset. Because they use information from multiple views, these fits are generally good and do not suffer from ambiguities that can plague pose estimation from single views. It is enticing to deploy this multi-view optimization approach towards our end-goal of estimating the pose and shape of all birds over time, but it is slow (initialization is usually far from the target) and requires multiple views in order to produce realistic poses. Nearly 40\% of the birds in our dataset were visible from one or fewer unoccluded views, however, indicating the need for a single-view approach. Second, from the multi-view fits, we extract distributions of shape and pose for birds in the aviary, which we use to create a synthetic dataset on which we train neural networks that regress pose and shape parameters from keypoints and silhouettes in a single view. Third, we train a second network to predict an instance segmentation and keypoints given a detection and corresponding bounding box.  Finally, we connect the keypoint and segmentation network to the pose regression network.  The full pipeline provides a pose and shape estimate from a single view image, which can be used to initialize further optimization (Figure \ref{fig:fullpipeline}).

\textbf{Bird detection in full images.}
We detect bird instances using a Mask R-CNN pretrained on COCO instance segmentation. We removed weights for non-bird classes (leaving bird and background) and then fine-tuned all layers on our dataset for 15 epochs in PyTorch.

\textbf{Keypoints and silhouette prediction.}
We train a convolutional neural network to predict keypoints and a silhouette given a detection and corresponding bounding box.  We modify the structure of High-Resolution Net (HRNet)~\cite{Sun2019}, which is state-of-the-art for keypoint localization in humans, so that it outputs masks in addition to keypoints. Our modified HRNet achieves 0.46 PCK@05, 0.64 PCK@10, and 0.78 IoU on our dataset.

\textbf{Skinned linear articulated bird model.}
To define an initial mesh, joint locations, and skinning weights, we used an animated 3D mesh of a bird model downloaded from the CGTrader Marketplace website. The model originally contained 18k vertices and 13k faces, but we removed vertices associated with body feathers, eyes, and other fine details to obtain a mesh with 3932 vertices, 5684 faces, and 25 skeletal joints (including a root joint, which is used for camera pose). We use the skinning weights defined in the original file.  In addition to skeletal joints, we define 12 mesh keypoints that correspond to the annotated semantic keypoints in our dataset. We obtain keypoint locations by identifying up to four mesh vertices associated with each keypoint and averaging their 3D locations.

\begin{figure}
\begin{center}
\includegraphics[width=0.75\linewidth]{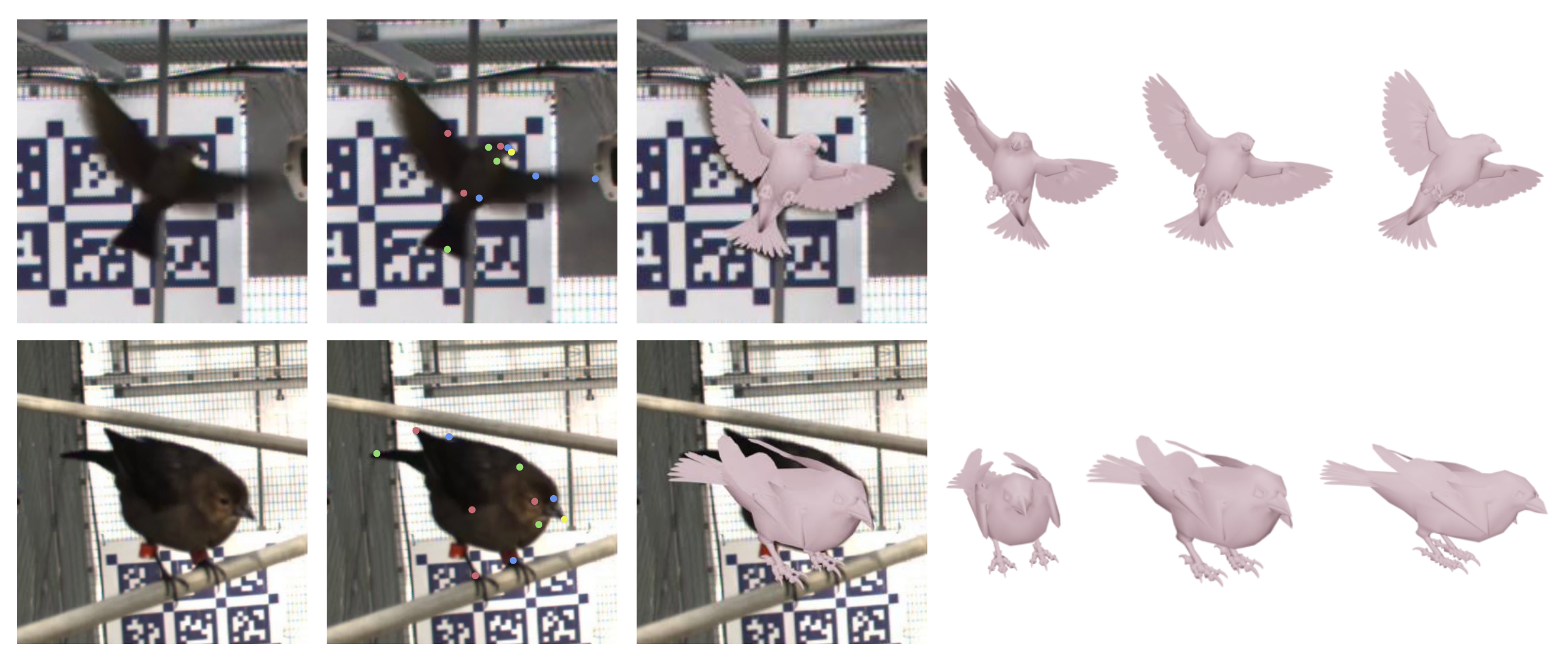}
\end{center}
   \caption{Our model is capable of capturing both perched and flying poses.}
\label{fig:short}
\end{figure}

To pose the model, we specify a function $M(\alpha, \theta, \gamma, \sigma)$ of bone length parameters $\alpha \in \mathbb{R}^J$ for $J$ joints, pose parameters $\theta \in \mathbb{R}^{3J}$ specifying relative rotation of the joints (and the rotation of the root relative to the global coordinate system) in axis-angle parameterization,  global translation inside the aviary $\gamma$, and scale $\sigma$, that returns a mesh $\mathcal{M} \in \mathbb{R}^{N \times 3}$, with $N=3932$ vertices. Unlike SMPL \cite{Loper2015} and SMAL \cite{Zuffi2017} models, we do not have access to 3D ground truth variation in shape, which prevents the use of shape coefficients drawn from a learned PCA shape space.  We mitigate this limitation by including an additional degree of freedom per joint, $\alpha_i$, that models the distance between parent and child joints, thereby capturing variation in the relative length proportions of the body and limb segments. When birds perch, their wings fold in on themselves and we found that this large deformation is not well modeled by LBS of a single bird mesh model (it is also difficult to capture and register in 3D scans).  To overcome this limitation, we use two template poses with identical mesh topology, bones, skinning weights, and keypoints, but with different initial postures: one for birds with their wings outstretched and another for birds with their wings folded (Figure~\ref{fig:short}). Finally, we also include an overall scale parameter to allow for consistent 3D multi-view estimation among cameras.

To form the mesh into a given pose, we modify the approach used in SMPL \cite{Loper2015} and SMPLify \cite{Bogo2016} to allow variable bone lengths. Starting with a template mesh $\mathcal{M}^T$ in a canonical pose with joint locations $\mathcal{J} \in \mathbb{R}^{J \times 3}$, we first calculate the position of each joint $i$ relative to its parent as
\begin{equation}
\mathcal{J}^o_i = \mathcal{J}_i - \mathcal{J}_{\mathrm{parent}(i)}.
\end{equation}
We then multiply this vector by $\alpha_i$ to adjust the distance between the two joints and form a new skeletal shape $\mathcal{J}'$, still in the canonical pose, with joint locations
\begin{equation}
\mathcal{J}'_i = \alpha_i \, \mathcal{J}^o_i + \sum_{j \in A(i)} \alpha_j \, \mathcal{J}^o_j,
\end{equation}
where $A(i)$ is the ordered set of joint ancestors of joint $i$ (i.e. all joints encountered moving along the kinematic tree from joint $i$ to the root). Finally, $\mathcal{J}' = J(\alpha)$ is transformed into the final pose using the global rigid transformation $R_\theta(\cdot)$ defined by pose and root orientation parameters $\theta$, and a LBS function $W(\cdot; \mathcal{M}^T)$ is applied. The final mesh vertices are
\begin{equation}
\mathcal{M} = M(\alpha, \theta, \gamma, \sigma) \stackrel{\mathrm{def}}{=} \sigma W(R_{\theta}(J(\alpha)); \mathcal{M}^T) + \gamma.
\end{equation}
The positions of 3D keypoints are calculated as $P(M(\alpha, \theta, \gamma))$, where $P(\mathcal{M}): \mathbb{R}^{N \times 3} \mapsto \mathbb{R}^{K \times 3}$ and $K$ is the number of keypoints. In practice $P$ is simply the average of four selected mesh vertices for each semantic keypoint.

\textbf{Optimization.}
To fit our bird model to detected keypoints, we introduce a fitting procedure similar to \mbox{SMPLify}, an optimization-based approach originally described by Bogo et al. \cite{Bogo2016}. Unlike SMPLify, we capture among individual variation using bone length parameters rather than body shape parameters and we fit to semantic keypoints rather than joint locations.  We minimize an objective function with a keypoint reprojection error term and silhouette error term for each camera $i$, two pose priors, and a prior on the relative 3D distances between joints. Specifically, we minimize:
\begin{equation}
E(\alpha, \theta, \gamma) = \sum_{\mathrm{cam} \, i} E_{kp}^{(i)}(\cdot; \cdot) + E_{msk}^{(i)}(\cdot; \cdot) + \lambda_\theta E_\theta(\theta) + \lambda_p E_p(\theta) + \lambda_b E_b(\alpha)
\label{eq:smplify_loss}
\end{equation}
with
\begin{equation}
E_{kp}^{(i)}(\alpha, \theta, \gamma; K_i, R_i, t_i, \mathcal{P}_i) = \sum_{\mathrm{kpt} \, k} w_k \rho(\parallel\Pi_{K_i, R_i, t_i}(P(M(\alpha,\theta,\gamma))_k - \mathcal{P}_{i,k}\parallel_2))
\label{eq:keypoint_loss}
\end{equation}
and
\begin{equation}
E_{msk}^{(i)}(\alpha, \theta, \gamma; K_i, R_i, t_i, \mathcal{S}_i) = \lambda_{msk} \parallel \mathcal{R}_{K_i, R_i, t_i}(M(\alpha,\theta,\gamma)) - \mathcal{S}_{i}\parallel_2.
\label{eq:mask_loss}
\end{equation}
Equation \ref{eq:keypoint_loss} is a weighted reprojection penalty (using the robust Geman-McClure function $\rho$~\cite{Geman1987}) between keypoints $\mathcal{P}_i$ and the projected mesh keypoints \sloppy $\Pi_{K_i, R_i, t_i}(P(M(\alpha,\theta, \gamma)))$ for pinhole projection function $\Pi(x) = K[R|t]\,x$. The bone lengths, $\alpha$, are the distances between parent and child joints, $\theta$ are the pose parameters, $\gamma$ is the translation in the global reference frame, $K_i$, $R_i$, and $t_i$ are the intrinsics, rotation, and translation, respectively, used in perspective projection for camera $i$, and $\mathcal{P}_i$ are the detected or annotated 2D keypoint locations in the image. Equation \ref{eq:mask_loss} penalizes differences between an annotated mask $\mathcal{S}_i$ and a rendered silhouette $\mathcal{R}_{K_i, R_i, t_i}(M(\alpha,\theta, \gamma))$ obtained using Neural Mesh Renderer \cite{Kato2018}. ${E_\theta(\theta) = |\theta - \theta_o|}$ is a pose prior that penalizes the $L_1$ distance from the canonical pose $\theta_o$. ${E_p(\theta) = \max(0, \theta - \theta_\mathrm{max}) + \max(0, \theta_\mathrm{min} - \theta)}$ linearly penalizes joint angles outside defined limits $\theta_\mathrm{min}$ and $\theta_\mathrm{max}$ and ${E_b(\alpha) = \max(0, \alpha - \alpha_\mathrm{max}) + \max(0, \alpha_\mathrm{min} - \alpha)}$ penalizes bone lengths outside limits $\alpha_\mathrm{min}$ and $\alpha_\mathrm{max}$. In the single-view setting, the pose prior ($E_\theta$) and joint angle ($E_p$) and bone length ($E_b$) limit losses are disabled and we use the Mahalanobis distance to the distribution of multi-view pose and shape estimates instead. We minimize the objective in~\ref{eq:smplify_loss} using Adam~\cite{Kingma2014} in PyTorch.

\textbf{Synthetic data and pose and shape regression.}
After performing multi-view optimization on 140 3D bird instances in our annotated dataset, we fit a multivariate Gaussian to the estimated pose parameters (pose, viewpoint, and translation). We then sample 100 random points from this distribution for each bird instance, project the corresponding model's visible keypoints onto the camera and render the silhouette, generating 14,000 synthetic instances for training. We keep the bone lengths of the original 140 instances, but add in random noise to the bone lengths for each sample.

We train pose and shape regression networks on the 14,000 synthetic single-view instances supervised by the ground truth pose and shape parameters. For the pose regression network inputs are 2D joint locations and targets are 3D rotations, which are first transformed to the representation proposed by Zhou et al. \cite{Zhou2019} before computing the $L^2$ loss. The pose regression network is an MLP with two fully connected layers with the final layer outputting $25*6+3$ translation parameters. The shape regression network takes in a mask and contains one $5 \times 5$ convolutional layer followed by four $3 \times 3$ convolutional layers and a fully connected layer with 24 outputs, corresponding to the 24 bone lengths. Each convolutional layer is followed by batch normalization and max-pooling layers. Training was performed for 20 epochs using Adam.

\section{The cowbird dataset}

\textbf{Image acquisition and aviary details.}
We captured video of 15 individual cowbirds (\textit{Molothrus ater}) in an outdoor aviary from March to June using eight synchronized cameras recording $1920 \times 1200$ images at 40 Hz.  The aviary is 2.5 meters in height and width and is 6 meters long. Cameras were positioned in the corners and oriented so that their combined fields view provided maximal coverage of the aviary volume by least four cameras. Intrinsic parameters were estimated for each camera using a standard checkerboard and the camera calibration package in ROS.  Extrinsic parameters for camera orientation and translation were estimated online via the TagSLAM package \cite{pfrommer2019tagslam} using arrays of fiducial markers permanently attached to the aviary walls.

\textbf{Dataset annotation and statistics.}
From the above recordings, we exported sets of synchronous frames from 125 ``moments" (1000 images) drawn from 10 days uniformly distributed over the recording period (an average of 12.5 uniformly distributed moments each day). On all images, we exhaustively annotated instance segmentation masks for all visible birds, producing over 6355 masks and bounding boxes. On a subset of 18 moments across six of the 10 days we also annotated the locations of 12 semantic keypoints on a total of 1031 masks (Figure ~\ref{fig:teaser}). We annotated the bill tip, right and left eyes, neck, nape, right and left wrists, right and left wing tips, right and left feet, and the tail tip. Statistics on the visibility of keypoints (Table S7) and a comparison with other animal datasets (Tables S4, S5) are in the supplementary material.

We manually associated keypoint annotations within each moment across camera views to create 3D instance ID tags. From the 3D instance ID tags, 64\%, 26\%, and 10\% of birds were fully or partially visible from four or more cameras, three cameras, and two or fewer cameras, respectively (Supplementary Table S6).  The average width $\times$ height of bird masks was $68 \times 75$ pixels (or $\approx$ 5\% of image width; the 5th and 95th percentiles of bird max dimensions were $17 \times 19$ and $239 \times 271$ pixels, respectively). We provide four types of test/train splits: by moment, by day, by time of day (morning vs. afternoon), and by season (March and April vs May and June).  Birds wore colored bands on their legs that, when visible, could provide the true ID of the bird, but we leave the potential application of this dataset to the Re-ID task for future work.

\section{Experiments}

\begin{figure}
\begin{center}
\includegraphics[width=0.95\linewidth]{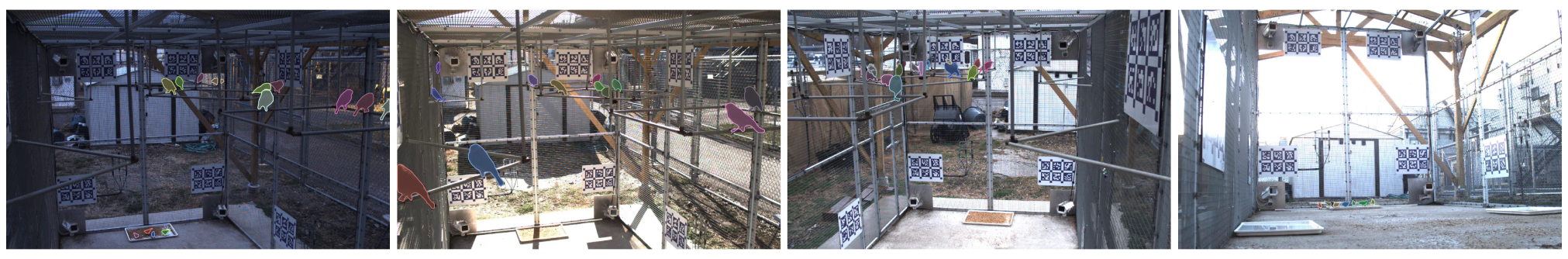}
\end{center}
   \caption{Instance detections made by a fine tuned Mask R-CNN network over a large range of lighting conditions and views. Best viewed in color.}
\label{fig:detections}
\end{figure}

\textbf{Detection.}
We first evaluate the performance of Mask R-CNN on instance segmentation of birds using our dataset. We show excellent generalization (AP = 0.52) when predicting masks on unseen days in the test set (Figure~\ref{fig:detections}). Further analyses and performance on additional splits of the dataset (e.g. split by time of day or season) are provided in Supplementary Table S1.

\begin{figure}
\begin{center}
\includegraphics[width=0.9\linewidth]{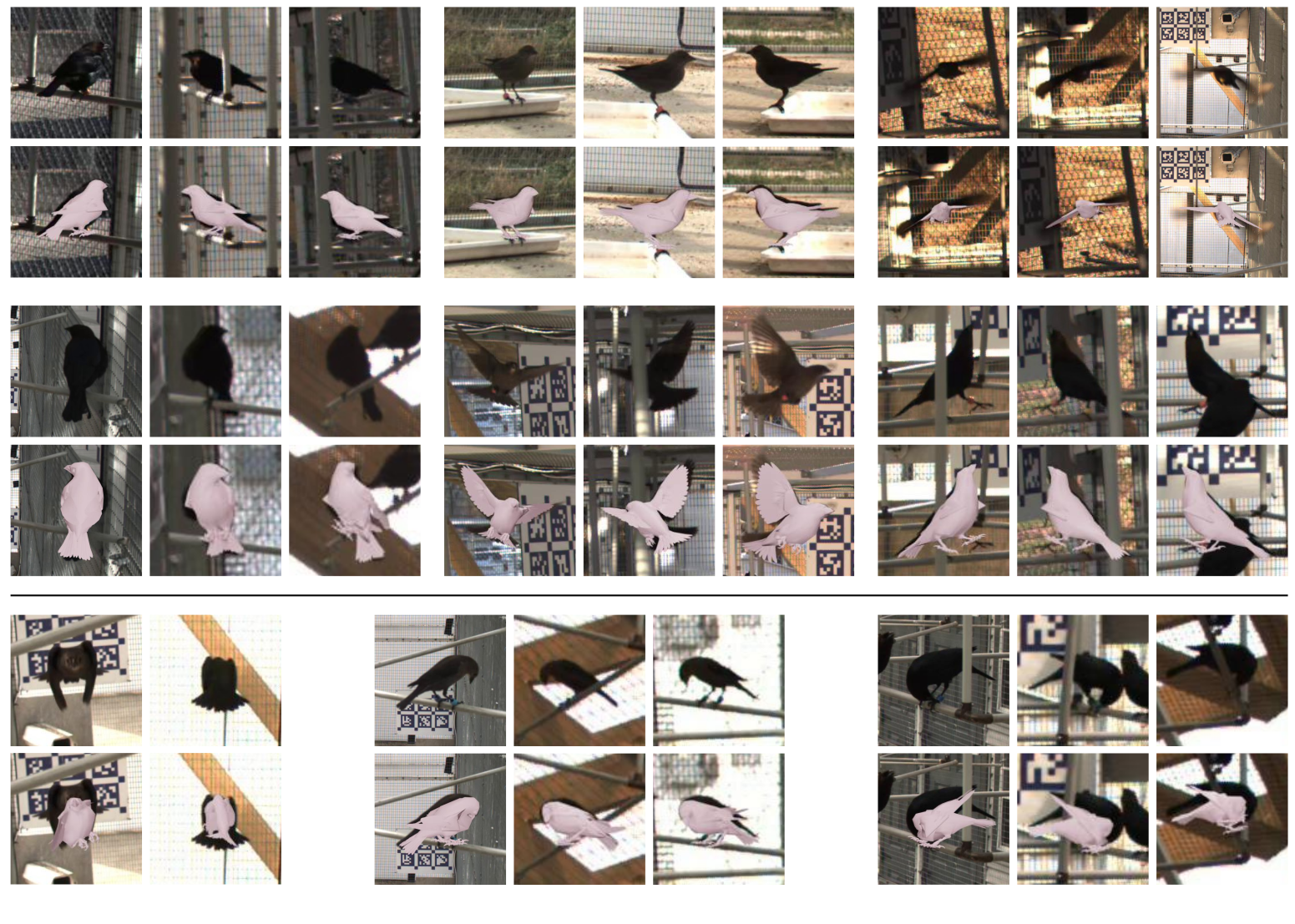}
\end{center}
   \caption{Multi-view optimization-based fits of the bird mesh to keypoint and mask annotations in our dataset (upper section). Failure cases are shown in the lower section.}
\label{fig:qualresults}
\end{figure}

\textbf{Multi-view optimization.}
We fit our articulated avian mesh model to annotations corresponding to each 3D bird instance in our keypoint dataset.  We fit using all keypoint labels from all available views.  We present qualitative results in Figure~\ref{fig:qualresults}.  Our fitting procedure resulted in many plausible results but also in many failure cases, shown in the bottom row of Figure~\ref{fig:qualresults}.  From the multi-view fits, we obtained a pose and shape space for the mesh model, which we display in the supplementary video.
\begin{figure}
\begin{center}
\includegraphics[width=0.8\linewidth]{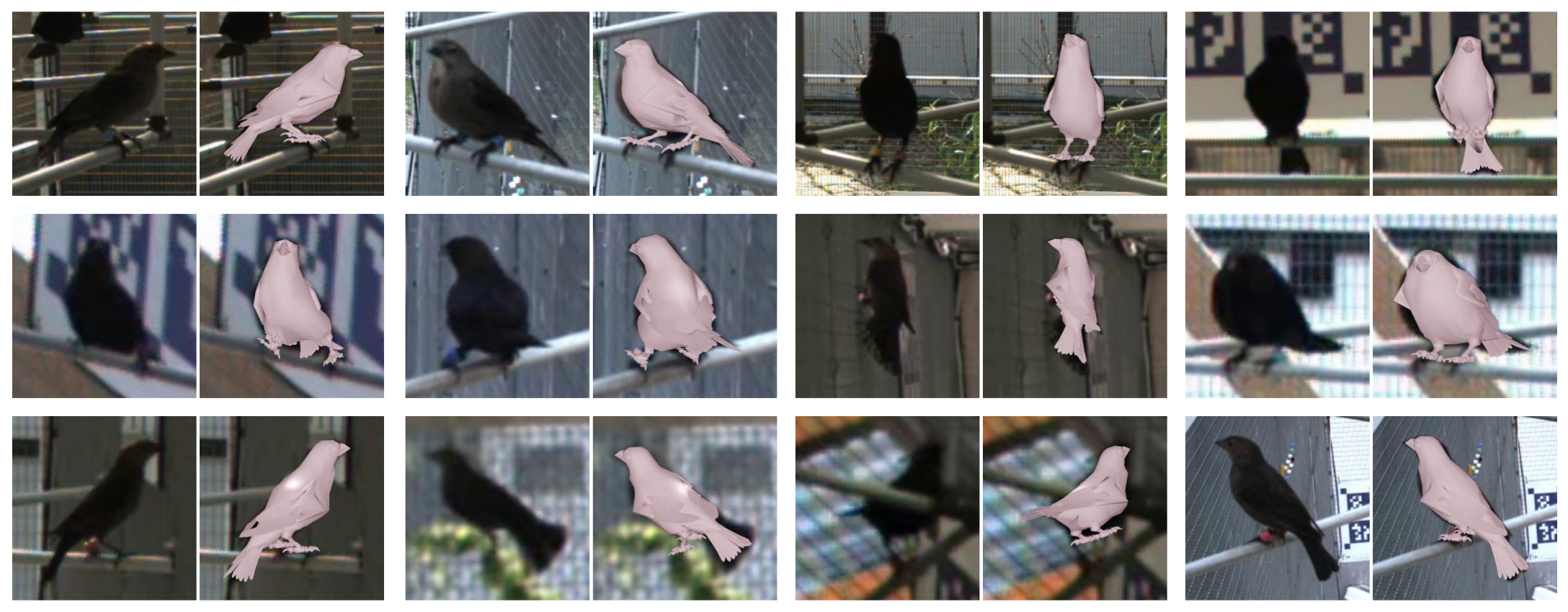}
\end{center}
   \caption{Regression-based recovery of bird pose and shape from a single view. Each panel shows the input image and refined mesh (see Figure \ref{fig:fullpipeline}).}
\label{fig:pipelineresults}
\end{figure}
We perform an ablation experiment to investigate the effects of pose priors and joint and bone limits on performance in the single-view setting. For each ablation, we remove the corresponding term from the objective and report its effect on the accuracy of projected mesh keypoints and silhouettes, which we report in Supplementary Table S3. We measure keypoint accuracy relative to ground truth keypoints using PCK at two thresholds calculated based on the largest dimension of the bounding box and we measure the accuracy of the projected silhouettes using IoU with ground truth masks.  We budget 500 iterations for fitting each instance for all settings.  The PCK increased as we removed the pose prior and bone limit (but not pose limit) terms from our objective.  This increase indicates the model is achieving a better fit to the keypoints, potentially at the cost of producing an unrealistic fit, as might be indicated by the simultaneous decrease in IoU as priors are removed.

\textbf{Do silhouettes improve multi-view optimization?}
We compared fits of the model with and without the silhouette term (Equation \ref{eq:mask_loss}) in the objective. The silhouette term improves IoU while only slightly affecting keypoint error (Table \ref{tab:maskablation}).  More importantly, the silhouette term allows the model to better capture changes in shape produced during feather puffing (Figure \ref{fig:wingstrokes}).

\begin{table}
\centering
\caption{Ablation study of the silhouette term in the multi-view optimization setting. PCK@05 and PCK@10 denote percent correct keypoints within 5\% and 10\% of bounding box width, respectively. Silhouettes improve IoU with minimal effect on keypoint error.}
\tabcolsep=0.85mm
\begin{tabular}{@{}lrccc}
\toprule
& weight ratio (kpt:mask) & PCK@05 & PCK@10 & IoU \\
\midrule
keypoints only   & N/A  & 0.356 & 0.631 & 0.540 \\
keypoints + mask & 10:1 & 0.355 & 0.637 & 0.560 \\
keypoints + mask & 1:1  & 0.328 & 0.618 & 0.624 \\
\bottomrule
\end{tabular}
\label{tab:maskablation}
\end{table}

\textbf{3D shape and pose recovery from a single view.}
Our single-view pipeline produces poses that are consistent across views (Table \ref{tab:crossview}, Supplementary Figure S1). To overcome scale ambiguity, we fix pose and shape and then find the Procrustes transformation (translation, rotation, and scaling) that minimizes keypoint reprojection error in each additional view. We also perform experiments to evaluate the individual components of our full pipeline (Table \ref{tab:fullpipeline}). We first compare pose regression alone (i.e. not optimizing after regression), single-view optimization alone (i.e. not initialized by pose regression network), and the full pipeline. Although the regression network alone is less ``accurate'' than single-view optimization (Table \ref{tab:fullpipeline}), the pose regression network produces good estimates of global pose, which allows optimization to proceed much faster. Additional examples are shown in Figure \ref{fig:pipelineresults}. Finally, we demonstrate that our model and bone length formulation generalize to similar bird species in the CUB-200 dataset (Supplementary Figure S3).
\begin{table}
\parbox{.48\linewidth}{
\centering
\caption{Cross-view PCK and IoU of projected meshes from the single-view pipeline. Values are averaged across all views except the view used to obtain the mesh. Ground truth pipeline input means the keypoint and mask network predictions (Figure \ref{fig:fullpipeline}) are replaced by ground truth annotations.}
\tabcolsep=0.75mm
\begin{tabular}{@{}lrccc}
\toprule
Pipeline input & PCK@05 & PCK@10 & IoU \\
\midrule
predictions & 0.313 & 0.632 & 0.589 \\
ground truth & 0.332 & 0.635 & 0.586 \\
\bottomrule
\end{tabular}
\label{tab:crossview}
}
\hfill
\parbox{.48\linewidth}{
\centering
\caption{Same-view evalutation of the single-view pipeline and ablations. Regression and optimization are performed using keypoint and mask predictions and evaluated against ground truth. Additional results are presented in Supplementary Table S2.}
\tabcolsep=0.75mm
\begin{tabular}{@{}lccc}
\toprule
& PCK@05 & PCK@10 & IoU \\
\midrule
regression   & 0.104 & 0.318 & 0.483 \\
optimization & 0.331 & 0.575 & 0.641 \\
reg. + opt.  & 0.364 & 0.619 & 0.671 \\
\bottomrule
\end{tabular}
\label{tab:fullpipeline}
}
\end{table}

\textbf{Failure cases.} Occasional failures resulted in unnatural poses, which are shown in Supplementary Figure S2. To evaluate the cause of these failures, two annotators inspected the same random sample of 500 crops and rated their confidence in each bird's pose (confident, semi-confident, not-confident). They then rated the predicted keypoints as good or bad for all crops. Finally, they viewed the mesh fits and rated each as a success or failure. We found that 84\% of confident, 35\% of semi-confident, and 12\% of not-confident crops were fit successfully. Bad keypoint detection was responsible for 60\% of failures. Even good fits are not perfect, particularly in the tail and feet. Adding more degrees of freedom to the model, such as tail fanning, and annotating additional keypoints on the toes would improve these areas.

\section{Conclusions}

We present an articulated 3D model that captures changes in pose and shape that have been difficult to model in birds. We provide a novel multi-view dataset with both instance masks and keypoints that contains challenging occlusions and variation in viewpoint and lighting. Our single-view pipeline recovers cross-view-consistent avian pose and shape, and enables robust pose estimation of birds interacting in a social context. We aim to deploy our pipeline in the aviary to better understand how individual interactions drive the formation of avian social networks.

An interesting feature of birds is that variation in a single individual's shape across time can be much larger than overall shape variation among individuals (e.g. due to feather fluffing shown in Figure \ref{fig:wingstrokes}). In the future, it will be interesting to apply our pipeline to video data and additional species to develop a more nuanced model of how shape varies across time, individuals, and species.

Capturing 3D pose is critical to understanding human and animal health and behavior.  Pose data produced by our pipeline will be useful for addressing how flying animals maneuver, negotiate cluttered environments, and make decisions while foraging or searching for mates, and how the collective behavior of a group arises from individual decisions.\\
\\
\textbf{Acknowledgements.} We thank the diligent annotators in the Schmidt Lab, Kenneth Chaney for compute resources, and Stephen Phillips for helpful discussions. We gratefully acknowledge support through the following grants: NSF-IOS-1557499, NSF-IIS-1703319, NSF MRI 1626008, NSF TRIPODS 1934960.

\clearpage
%
%
\bibliographystyle{splncs04}
\bibliography{egbib}
\end{document}